\documentclass{article}

\usepackage{arxiv}

\usepackage[utf8]{inputenc} 
\usepackage[T1]{fontenc}    
\usepackage{hyperref}       
\usepackage{url}            
\usepackage{booktabs}       
\usepackage{amsfonts}       
\usepackage{nicefrac}       
\usepackage{microtype}      
\usepackage{lipsum}
\usepackage{hyperref}    
\usepackage{amsfonts}    
\usepackage{amsmath}
\usepackage{amssymb}
\usepackage[english]{babel}
\usepackage[square,numbers]{natbib}
\usepackage{graphicx}
\usepackage{xcolor}
\usepackage{rotating}
\usepackage{pdflscape}
\usepackage{float}

\title{Symptom extraction from the narratives of personal experiences with COVID-19 on Reddit}

\author{
  Curtis Murray \\
  The University of Adelaide\\
  Adelaide, SA 5005 \\
  \texttt{curtis.murray@adelaide.edu.au} \\
   \And
 Lewis Mitchell \\
  The University of Adelaide\\
  Adelaide, SA 5005 \\
  \texttt{lewis.mitchell@adelaide.edu.au} \\
  \And
 Jonathan Tuke \\
  The University of Adelaide\\
  Adelaide, SA 5005 \\
  \texttt{simon.tuke@adelaide.edu.au} \\
  \And
 Mark Mackay \\
  The University of South Australia\\
  Adelaide, SA 5000 \\
  \texttt{mark.mackay@unisa.edu.au} \\
}

\begin{document}
\maketitle

\begin{abstract}
Social media discussion of COVID-19 provides a rich source of information into how the virus affects people's lives that is qualitatively different from traditional public health datasets. 
In particular, when individuals self-report their experiences over the course of the virus on social media, it can allow for identification of the emotions each stage of symptoms engenders in the patient.
Posts to the Reddit forum r/COVID19Positive contain first-hand accounts from COVID-19 positive patients, giving insight into personal struggles with the virus. 
These posts often feature a temporal structure indicating the number of days after developing symptoms the text refers to. 
Using topic modelling and sentiment analysis, we quantify the change in discussion of COVID-19 throughout individuals' experiences for the first 14 days since symptom onset. 
Discourse on early symptoms such as fever, cough, and sore throat was concentrated towards the beginning of the posts, while language indicating breathing issues peaked around ten days. 
Some conversation around critical cases was also identified and appeared at a roughly constant rate.
We identified two clear clusters of positive and negative emotions associated with the evolution of these symptoms and mapped their relationships.
Our results provide a perspective on the patient experience of COVID-19 that complements other medical data streams and can potentially reveal when mental health issues might appear.
\end{abstract}


\section{Introduction}

The COVID-19 pandemic is having an unprecedented impact on the world, in ways that are yet to be fully understood. 
As of May 12th, 2020, there have been 4,013,718 confirmed cases, with 278,993 confirmed deaths according to the World Health Organization\footnote{\url{https://www.who.int/emergencies/diseases/novel-coronavirus-2019}}. 
As a result, extensive discourse on various aspects of COVID-19 has emerged on social media \cite{ordun2020exploratory}. 
This discourse demonstrates how individuals react to the virus and how it affects them. 
Analysing the social media response to COVID-19 allows us to better understand its effect on populations, and potentially make informed decisions on how to respond to this, and future pandemics.

Recent work in Natural Language Processing (NLP) has revealed various dynamics of the social media response to COVID-19. 
The majority of this work focuses on the discussion of COVID-19 on Twitter.
\citet{sha2020dynamic} use topic modelling and network analysis to understand how decision-makers influence each other and identify leaders for different topics. 
\citet{chen2020eyes} use sentiment analysis to indicate there may be racism behind controversial terms such as ``Chinese Virus'', as these coincide with anger and other negative emotions.
\citet{li2020we} use sentiment analysis to analyse tweets, showing how COVID-19 affects mental health.
\citet{wicke2020framing} shows how war-related terminology is commonly used to frame the discourse on COVID-19.

However, little work has been conducted using social media discourse to consider the personal impact of being COVID-19 positive. 
In particular, in cases where individuals use social media to diarise their personal experiences after being diagnosed with the virus, it should be possible to extract ``emotional arcs'' charting their average emotional state over time \cite{reagan2016emotional}.
Social media data has previously been used to track population-level health measures such as obesity \cite{alajajian2017lexicocalorimeter}, as well as mental health issues such as depression \cite{reece2017forecasting} and body image \cite{tiggemann2018tweeting}. 
These can be beneficial to health care providers, enabling early warning of potential mental health issues, as well as a general overview of the patient experience.


The idea that social media has a role to play in transforming the health sector is not new and was raised more than a decade ago \cite{hawn2009take}. Apart from service delivery, social media has been used and has potential to be used as an engagement mechanism for health policy development \cite{o2017using}.

Despite the health care sector being yet to widely adopt social media as a part of a business or policy strategy, it provides an opportunity to better engage with a range of stakeholders, including policymakers \cite{charalambous2019social}. 
Increasing adoption, however, needs to be supported by evidence of benefit.  
Social media is a rapidly maturing communication medium that has potential to quickly inform patients of policy and evidence decisions \cite{roland2018social}. 
The potential to misunderstand the limitations of social media in information sharing, however, may not be recognised.

One benefit of using social media is that it provides an opportunity for more stakeholders’ voices to be heard, and therefore influence policy. A recent study which undertook a review of how media influences health policy determined there was insufficient evidence to guide decision-makers on how to use media when in developing health policy \cite{bou2017using}. While the focus of the review was limited to more traditional media, there is no reason to believe the findings do not also apply to the use of social media.

To date it would seem much of the focus of the discussion about using social media to influence health policy has been centred around information sharing and engagement \cite{charalambous2019social, bou2017using}. The COVID-19 pandemic has seen a need for rapid decision-making. Much has been made of various modelling to identify the potential spread of the disease and also the impact social distancing and other more innovative mechanisms will have on controlling outbreaks \cite{shaw2020governance}. However, the ability to collect non-traditional data arising from social media relating to the pandemic appears to have been largely overlooked as a response to influence health policy.
Here, we attempt to address this gap, through using social media data from Reddit to better understand the patient experience through the stages of COVID-19 illness.

\section{Data}

Posts to the Reddit community \texttt{r/COVID19Positive}\footnote{\url{https://www.reddit.com/r/COVID19positive/}} contain detailed recounts of personal experiences with COVID-19. The authors give the posts labels, or ``flairs'', that represent their relationship with COVID-19.
Of particular interest in this work are the flairs reserved for those who have tested positive, ``Tested Positive - Me'' and ``Tested Positive''. 
Often, these posts are in the form of daily journal entries, where references to specific days of the author's experiences after developing symptoms are made. 
We exploit this structure to annotate the text within posts having relevant flairs by this reference date using regular expressions. 
As this annotation is relative to the onset of symptoms for each infectious patient, grouping by day number can then reveal common themes in discourse at particular stages in experiences with COVID-19. 
We perform topic modelling to reveal common themes in the symptoms described at different stages of illness,
and sentiment analysis to reveal how patients feel during each of these stages.
By correlating the trends of each over time, we connect patients' descriptions of their symptoms to their emotions. 
Using this information in conjunction with traditional evidence-based medical studies provides a different understanding of the stages of COVID-19 illness, hopefully giving medical practitioners and policymakers new tools to respond to this ongoing crisis.

A total of 4,610 posts to r/COVID19Positive were accessed using the Pushshift API\footnote{\url{https://github.com/pushshift/api}} from March 14th, 2020 to May 12th, 2020. To obtain first-hand accounts of users' experiences with COVID-19, we filter these posts to the 609 non-empty posts that have the flair ``Tested Positive - Me'' or ``Tested Positive''.

\subsection{Data Preprocessing}

\begin{figure}[h!]
 \centering
 \includegraphics[width=1\textwidth]{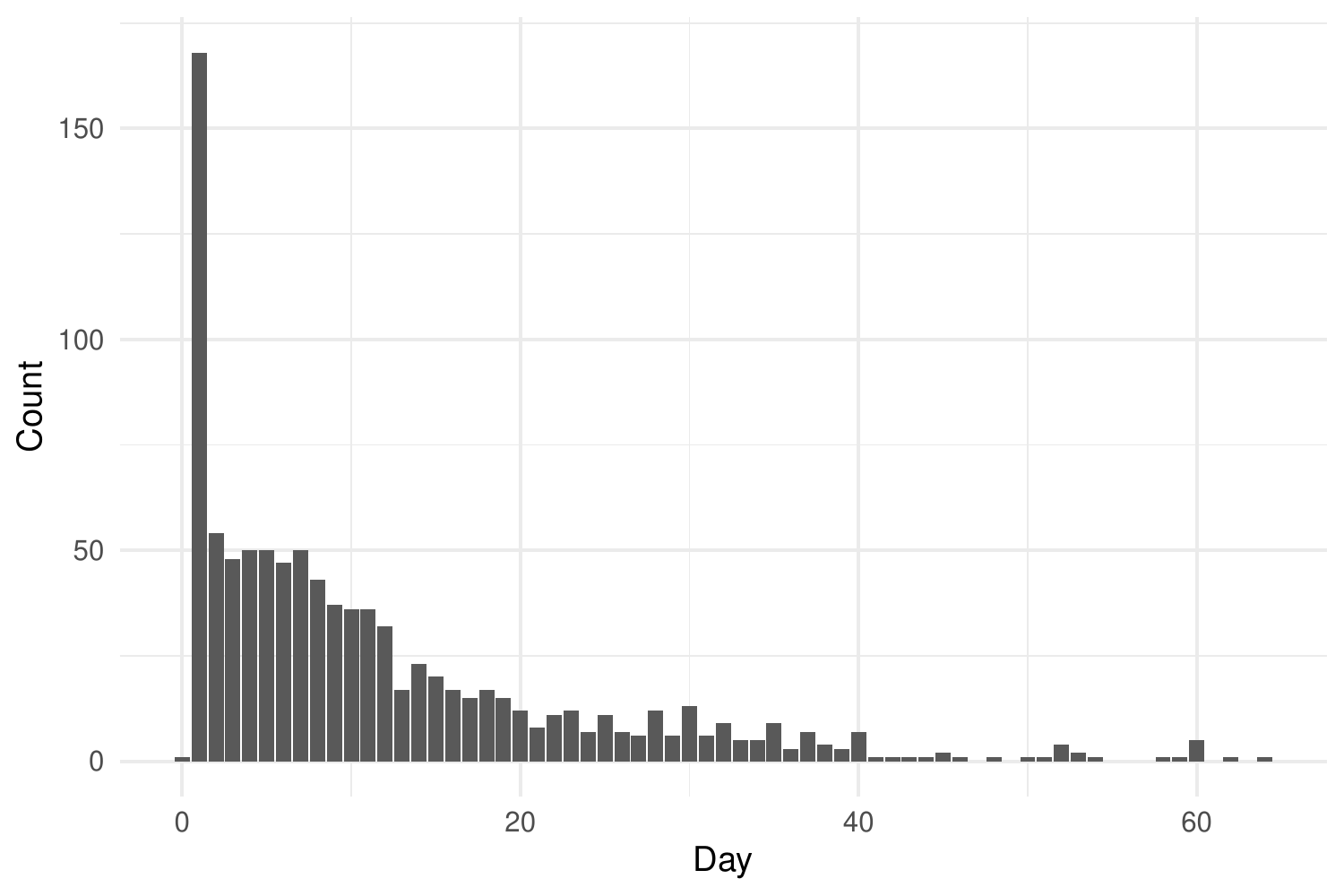}
 \caption{Bar chart showing mentions of specific number of days after symptom onset. One post was omitted on Day 150.}
  \label{fig:plot_day_counts}
\end{figure}

Annotating text within the \texttt{r/COVID19Positive} posts by the number of days since the author developed symptoms provides a collective timeline of the patient experiences. It was observed that journal-formatted posts fall into two main categories for describing dates, using the format ``Day $x$'' to describe the number of days since testing positive, or with real dates. We make the assumption that any post follows only one of these formatting rules. In the latter case we set ``Day 1'' to be the first date mentioned, and as follows for subsequent dates. Before there is a reference to a day or date, we annotate the text by Day: NA. An exception to this is made when a title mentions a specific day. If it does, we instead annotate this text by that day. If neither the ``Day $x$'' format, nor the absolute date format is detected, we discard the post as it does not contribute to our understanding of the collective timeline. Table \ref{tab:types} summarises the number of posts meeting these formats. The number of annotations for the number of days since developing symptoms is displayed in Figure \ref{fig:plot_day_counts}.  

\begin{table}[h!]
\centering
\caption{Number of posts adhering to each format.}
\label{tab:types}
\begin{tabular}{lr}
\toprule
Format & Count\\
\midrule
Daily Journal & 166\\
Absolute Date & 126\\
None & 317\\
\bottomrule
\end{tabular}
\end{table}

A considerable number of posts (57.5\%) mention the first day of symptom onset (Figure \ref{fig:plot_day_counts}). A sharp decrease in the number of posts is observed for the second day (18.5\%). There is a small decrease in the number of posts after the first week, followed by a more rapid decrease in the number of posts over the next two weeks. By the third week, there are few posts for each day; however, the decrease from this point is slight, resulting in a heavy tail. After 40 days, the number of posts becomes negligible. We omit one post from Figure \ref{fig:plot_day_counts} that occurred at Day 150. In total, after preprocessing, there were 292 posts from 223 unique authors, averaging 1.31 posts per author. There were 966 mentions to specific days, with each user referencing 4.33 days on average. These mentions to specific days, as well as the preliminary text at the start of a post before a day mention (for which we annotate as Day: NA), split the posts into 1,179 documents for the purpose of our analysis.

We use each of the 1,179 documents obtained from the day annotation when topic modelling, however as the number of posts for each day after two weeks of being COVID-19 positive is small (below 25 each day), we focus our analysis on the first two weeks of posts only. This period a pragmatic choice, and because it coincides with the quarantine period of 14 days in the United States\footnote{\url{https://www.cdc.gov/coronavirus/2019-ncov/faq.html}}.

\section{Topic Modelling}


We use network topic modelling \cite{gerlach2018network} to model the COVID-19 discourse using \textit{topics}. 
The notion of a topic is familiar with our general conversational use, capturing words used together in similar contexts.
More formally, we define a topic to be a mixture of words, $p(\text{word} | \text{topic})$. 
Through topic modelling we can represent documents as a mixture of topics, $p(\text{topic} | \text{document})$. 
Latent Dirichlet Allocation (LDA) has been a common approach \cite{blei2003latent}, using a generative model that uncovers topics by considering the posterior probability of the topics, given the data.

Network topic modelling shows community detection in a bipartite document-word network achieves the same goal as LDA \cite{gerlach2018network},
using a hierarchical stochastic block model (hSBM) \cite{peixoto2014hierarchical,peixoto2015model,peixoto2017nonparametric} approach. The use of a hSBM in the context of topic modelling avoids some undesirable properties of LDA, and may elucidate more appropriate topics \cite{gerlach2018network}.

We used a hSBM, where each block of text referencing distinct days within a post comprised a document. 
The removal of stopwords\footnote{\url{https://rdrr.io/cran/tidytext/man/stop_words.html}} from our documents produced more meaningful topics. Numbers were not removed as they were often discussed in the context of having a fever.



\begin{figure}[h!]
 \centering
 \includegraphics[width=1\textwidth]{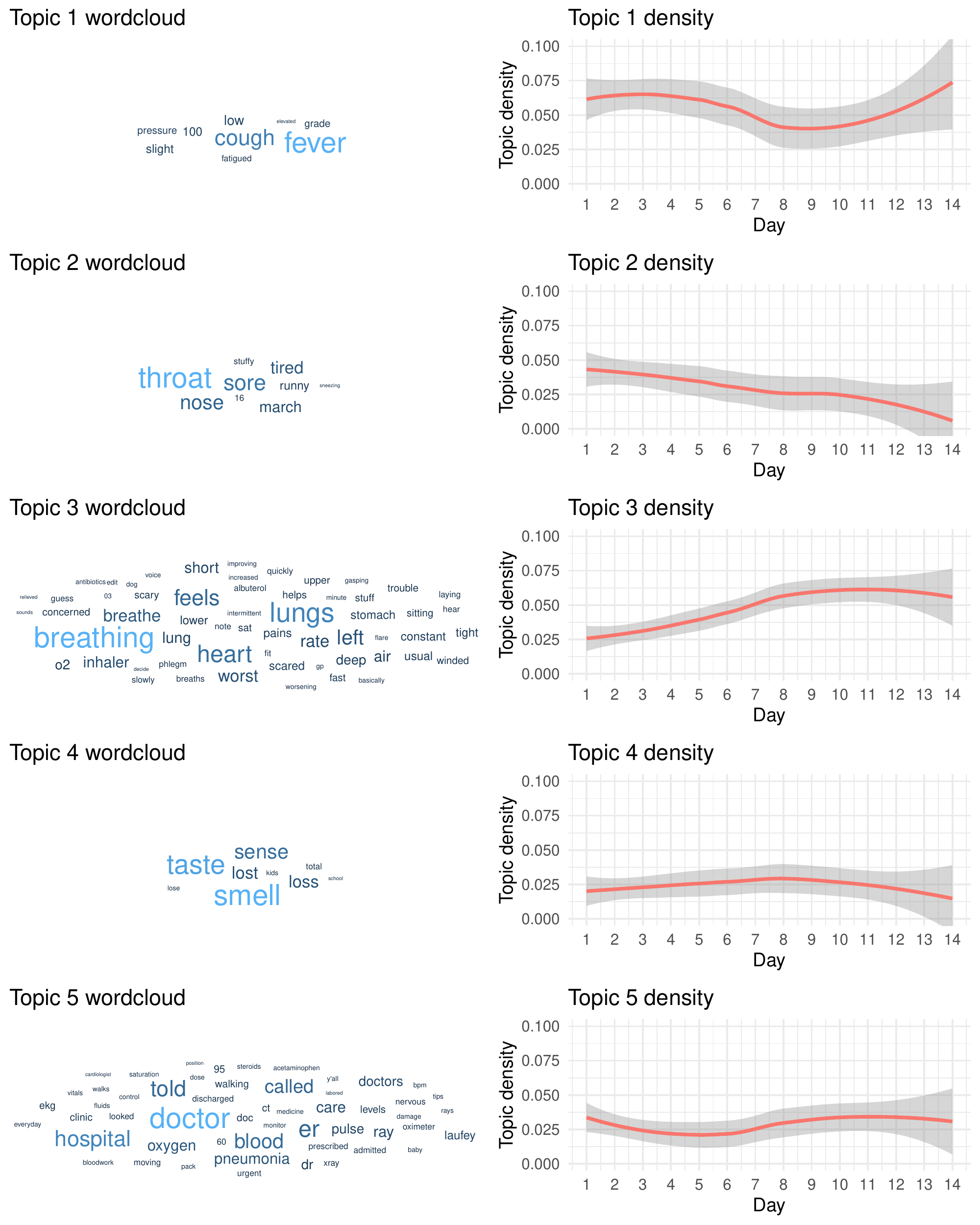}
 \caption{Select topics represented with word-clouds (left), with corresponding topic densities $p(\text{topic} | \text{document})$ (right) and LOESS smoothing.}
  \label{fig:merged_plots}
\end{figure}

Select topics found through topic modelling are represented with word clouds in Figure \ref{fig:merged_plots}, where the size of each word is proportional to the density $p(\text{word} | \text{topic})$. Alongside these word clouds, we plot the corresponding document topic densities $p(\text{topic} | \text{document})$, indexed by the number of days since first developing symptoms with a LOESS curve fitted. Presented are five topics selected from our topic model. A full display of all topics is available in the supplementary material online\footnote{\url{https://github.com/curtis-murray/COVID_Symptom_Extraction_Appendix}}. Notably, most topics appear to represent distinct groups of symptoms. 

Topic 1 captures conversation regarding fever, featuring common usage of the terms “fever”, “cough”, and “fatigued”. Fever is a near-ubiquitous symptom of COVID-19, found in  94.3\% of cases \cite{chen2020clinical}, and the most common symptom present at the onset of illness (87.1\%), followed by cough (36.5\%), and fatigue (15.7\%) \cite{chen2020clinical}. Moreover, fever is understood to be a precursory indicator of COVID-19, and hence likely prevalent in the discussion at the immediate onset of a case. The density of this topic, and all other topics, indicates what proportion of the discussion the topic features in at a point in the collective timeline. Results show an early peak in the discussion of this topic, which deteriorates after five days. A small drop in the level of discussion follows until Day nine, met by a resurgence in the level of discussion. This drop in discussion corresponds to a degree with the median duration of fever, which was reported as being 10 days \cite{chen2020clinical}. The resurgence in this topic density subsequent to this may potentially be a result of a worsening cough, as we do not expect a return of fever.

Topic 2 features the words ``throat'', ``nose'', accompanied by ``sore'' and ``runny''. These terms represent less common respiratory symptoms at onset \cite{chen2020clinical}. This topic exhibits an initial high density. In comparison to the fever topic, the density is slightly reduced, and exhibits immediate monotonic decay at a roughly constant rate. By 14 days, discussion around this topic almost entirely subsided.

Topic 3 features the characteristics of a more severe COVID-19 case, with terms indicating breathing and lung difficulties. This corresponds with our understanding of COVID-19 being a respiratory illness. This topic is initially rarely used, as COVID-19 does not immediately spread to the lower respiratory tract. After 10-11  days, discussion on this topic peaks and is used at a roughly consistent amount over the remainder of the duration.

Topic 4, interestingly, indicates conversation on anosmia and ageusia; the loss of ability to smell and taste respectively. As COVID-19 is a respiratory illness, it can affect the ability to smell and taste. One study supports this, finding 5.1\% of (214) hospitalised patients with COVID-19 in Wuhan, China was characterised with anosmia, and 5.6\% with ageusia \cite{mao2020neurological}.

Topic 5 corresponds with a discussion that may imply the condition has escalated to the level where the person has been admitted to hospital, based on the use of medical testing terminology. We see the density for this is relatively low throughout the timeline. The density is marginally larger for the first day of symptom onset. One cause for this may lie in testing for COVID-19, resulting in discourse around medical terminology. The small drop in density that follows over the next five days may be a result of going home and waiting it out. The increase thereafter may be attributed to the breathing difficulties identified in Topic 3, escalating the condition to hospital.

These results may not represent the proportions of discussion that feature any one particular symptom. 
A simple binary search for a particular word may be more appropriate for this task. 
Instead, this topic modelling allows for automatic extraction of themes in the narratives. 
There is no requirement for domain knowledge to be applied. 
A manual search, such as that mentioned above, may be to narrow and fail to capture other important aspects of the discussion, and overlook unexpected results.

\section{Sentiment Analysis}

\begin{figure}[h!]
 \centering
 \includegraphics[width=1\textwidth]{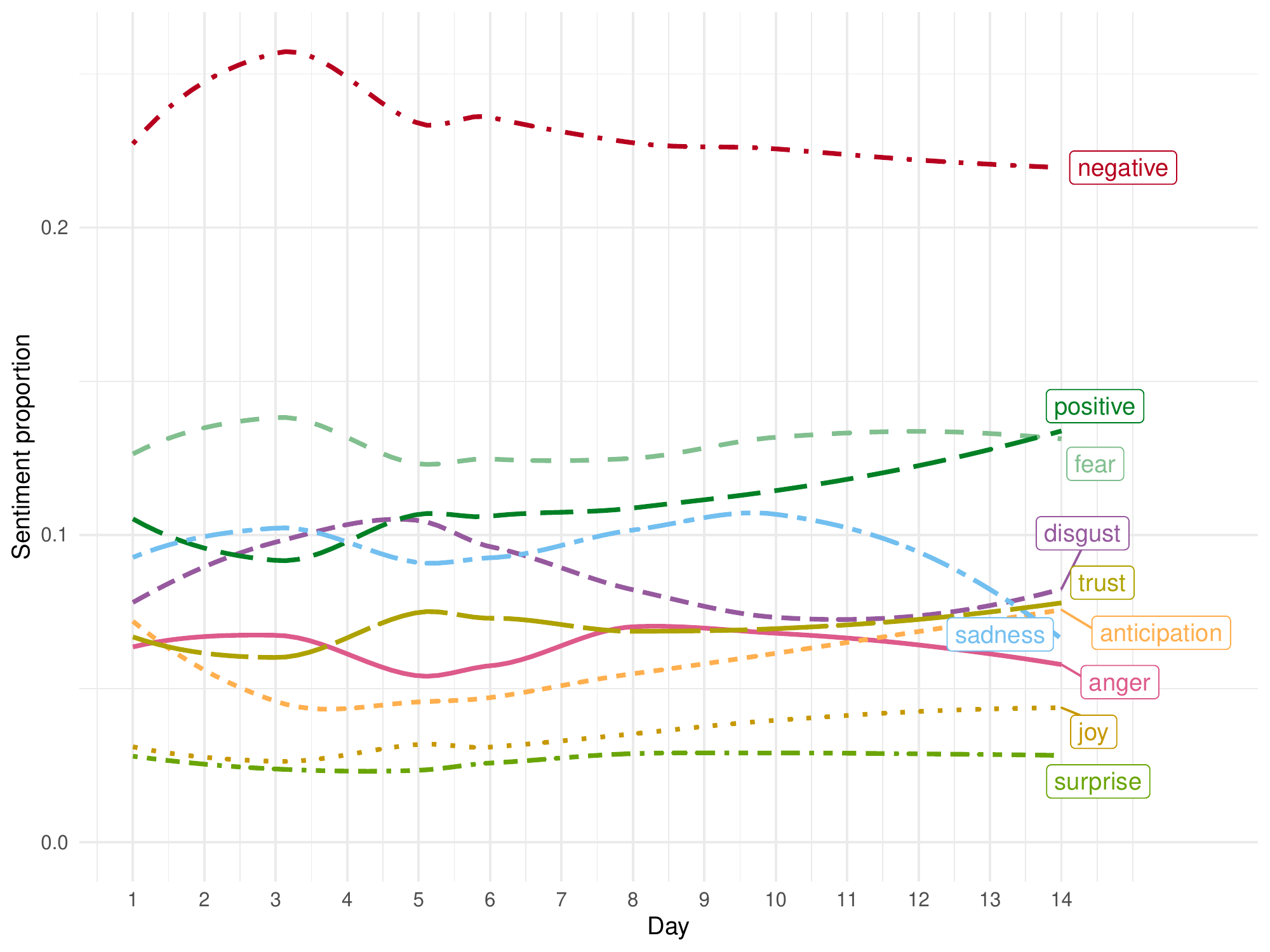}
 \caption{Plot showing the proportion of emotion carrying words for each sentiment group across the first 14 days of the COVID-19 collective timeline.}
  \label{fig:sentiment_nrc}
\end{figure}

Sentiment analysis using the NRC sentiment lexicon allows us to express each document as comprising 10 basic emotions: anger, anticipation, disgust, fear, joy, negative, positive, sadness, surprise, and trust \cite{mohammad13}. The top 10 most common terms used for each sentiment are shown in Table \ref{tab:sentiment_words}, where the colour of each sentiment is defined using Plutchik’s wheel of emotions \cite{plutchik1980emotion}. The results of applying this sentiment analysis to the posts to \texttt{r/COVID19Positive} marked with the tag ``Tested Positive - Me'' are displayed in Figure  \ref{fig:sentiment_nrc}.
The term ``feeling'' was removed from our corpus for the purpose of sentiment analysis as it appeared in all sentiment groups. Additionally, we remove the terms ``positive'' and ``negative'', as these terms were often use to express the result of a COVID-19 test, and do not bear positive or negative emotions as would otherwise be indicated by the NRC sentiment lexicon. 
We remark that the proportions calculated here for each emotion are proportions out of all emotion carrying words posted about each day.

\begin{table}[h!]
\centering
\caption{Top 10 most common terms used for each sentiment, ordered by prevalence.}
\label{tab:sentiment_words}
\begin{tabular}{ll}
\toprule
Sentiment & Terms\\
\midrule
\textcolor[HTML]{DD5A8B}{anger} & smell, sore, bad, anxiety, loss, hot, shit, painful, attack, hit\\
\textcolor[HTML]{FFAF4E}{anticipation} & time, pretty, anxiety, hope, start, finally, result, daily, coming, continue, develop\\
\textcolor[HTML]{96599E}{disgust} & cough, smell, bad, nose, sick, weird, nausea, finally, shit, stomach\\
\textcolor[HTML]{81BF8F}{fear} & fever, pain, hospital, worse, bad, anxiety, flu, loss, pneumonia, infection\\
\textcolor[HTML]{C79801}{joy} & pretty, food, hope, finally, lucky, safe, found, intense, weight, glad\\
\addlinespace
\textcolor[HTML]{B7001F}{negative} & cough, pain, smell, sore, headache, worse, bad, fatigue, sick, tired\\
\textcolor[HTML]{008025}{positive} & doctor, pretty, sense, food, hope, completely, nurse, eat, received, rest\\
\textcolor[HTML]{72BFF0}{sadness} & pain, sore, hospital, worse, bad, sick, anxiety, negative, loss, lost\\
\textcolor[HTML]{6AA507}{surprise} & hope, finally, lucky, leave, intense, mouth, suddenly, occasional, catch, guess, weight\\
\textcolor[HTML]{AEA200}{trust} & doctor, hospital, pretty, food, hope, nurse, finally, safe, experienced, usual\\
\bottomrule
\end{tabular}
\end{table}

Negativity is the dominant emotion exhibited in the posts, making up over 0.2 of the emotion carrying words. There is an increase in negativity to over 0.25 for the first three days, thereon gradually decaying to approximately 0.22. Positivity negatively correlates with negativity, increasing from 0.10 to 0.13 throughout the timeline, and dipping at roughly the same time as negativity peaks.

Fear is prevalent in COVID-19 discourse, being the second most common emotion on most days. 
The level of fear in posts is approximately constant throughout the first 14 days, with a small peak around the third day coincident with negativity. 
As shown in Table \ref{tab:sentiment_words}, this is likely due to the fact that the word ``fever'' is labelled with the ``fear'' emotion in NRC.

Another pervasive emotion is sadness, making up approximately 0.10 of the emotion carrying words for the first 11 days. After this point, its use reduces to 0.07 by Day 14. This may be an indicator of recovery -- as Table \ref{tab:sentiment_words} shows, the sadness emotion includes the word ``hospital''.

There is a notable increase in the proportion of disgust over the first five days, which aligns with early symptom onset. After this point, disgust drops off to a minimum at Day 11, with a small resurgence thereafter.

Anticipation levels drop considerably from day one to four, then increase monotonically over the remaining 14 days. Early anticipation may be attributed to wondering if they tested positive and how their condition will develop -- words like ``anxiety'' and ``time'' feature highly in the list of words associated with this emotion in Table \ref{tab:sentiment_words}.

Particularly uncommon emotions are joy and surprise. Joy has a substantial relative increase in density as the timeline progresses. This may be attributed to recovery.

\section{Sentiment-Topic Correlation}

\begin{figure}[h!]
 \centering
 \includegraphics[width=1\textwidth]{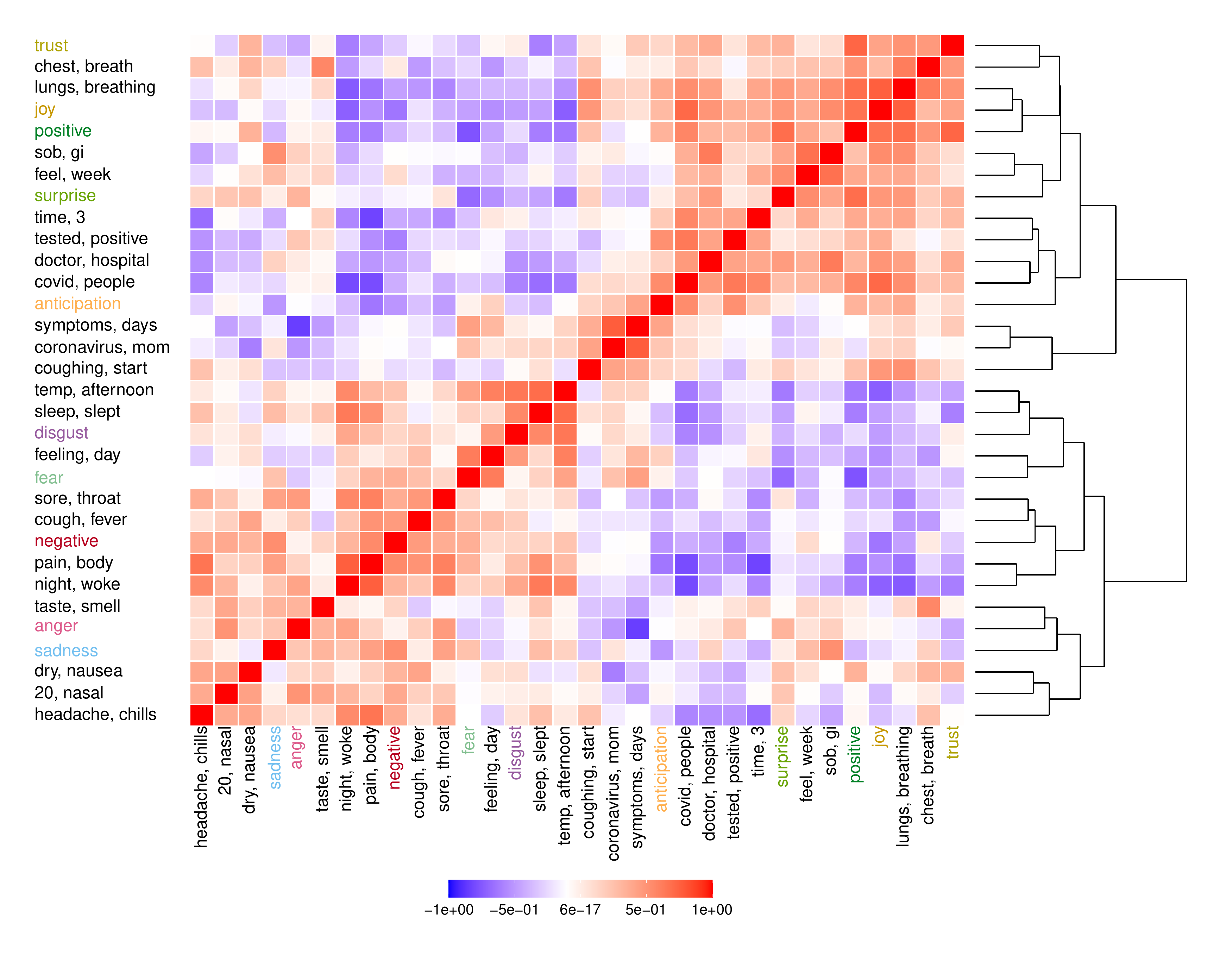}
 \caption{Heatmap of the Pearson correlations between topics and sentiments, arranged using hierarchical clustering.}
  \label{fig:heatmap}
\end{figure}

We now compare the trends in topic proportions over time with those for sentiment proportions, to reveal associations between the two. 
We use the Pearson correlation between the (mean) topic densities for each day, as well as the sentiment densities for each day to do this. 
The results are displayed as a heat-map in Figure \ref{fig:heatmap}, with structure highlighted by hierarchical clustering and a dendrogram. 
We represent each topic with its two most common words, and sentiments by the colours used previously.

This representation reveals two clear clusters of co-varying topics and emotions.
These broadly correspond to a ``positive'' cluster, which by inspection of the topics (see Figure \ref{fig:heatmap}) consists of topics and emotions which trend upwards over time,
and a ``negative'' cluster which peaks early on.

Each post, prior to topic modelling and sentiment analysis, is represented as a bag-of-words, which is a vector whose elements encode the number of occurrences of each word in that post. This is a high-dimensional representation of a post, where the dimension is equal to the number of unique words in the corpus. Both topic modelling and sentiment analysis can then be viewed as dimension reduction tools, taking the high dimensional bag-of-words representation to a low-dimensional topic-space $T$, and sentiment-space $S$ respectively. For a full comparison between topics and sentiments, we embed posts in the sentiment-topic-space $T \times S$, the Cartesian product of spaces. However, doing so results in a space whose dimension is the sum of the dimension of the topic-space and sentiment-space, and hence is still relatively high. We therefore use multidimensional scaling (MDS) to embed points from this high-dimensional space into a two-dimensional space. This is achieved by considering only the \textit{dissimilarity} between all points in the high-dimensional space, and not the points themselves. Points with small dissimilarity are placed close together, and those with substantial dissimilarities far apart.

\begin{landscape}
 \begin{figure}
 \centering
 \includegraphics[width=1.2\textwidth]{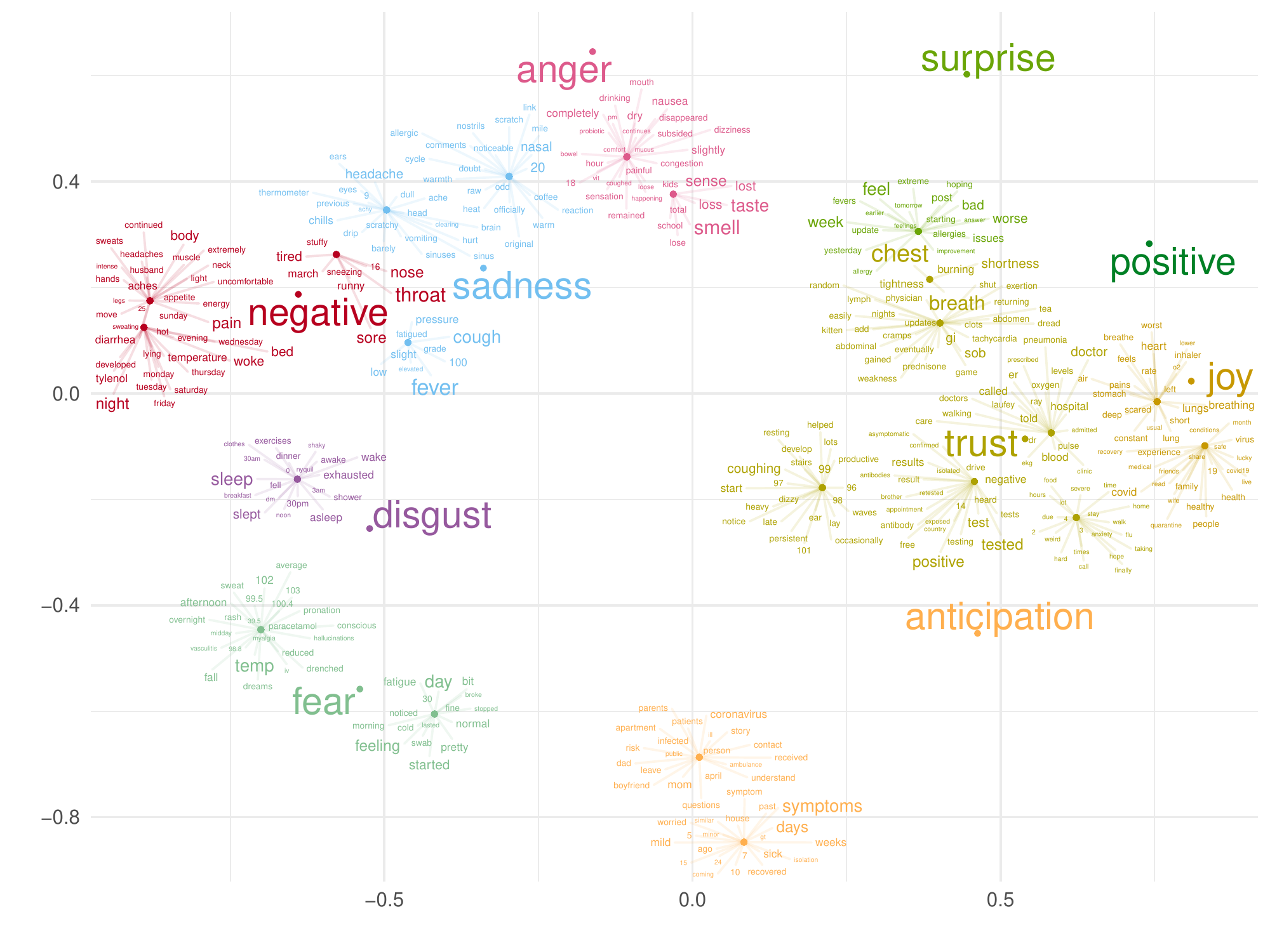}
 \caption{Two-dimensional representation of sentiment-topic-space using MDS with 1-$\rho_{X,Y}$ as a measure of dissimilarity. Topics are represented with word-clouds where the size of each word is related to the word-density $p(\text{word}|\text{topic})$. Topics are coloured by the colour of the nearest sentiment in the two-dimensional embedding.}
  \label{fig:MDS}
  \end{figure}
\end{landscape}

The Pearson correlations found above can be manipulated into a measure of dissimilarity by taking $d(X,Y) = 1-\rho_{X,Y}$. When $\rho_{X,Y} = 1$, i.e. $X$ and $Y$ are perfectly correlated, we find $d(X,Y) = 0$. Similarly, when $X$ and $Y$ are uncorrelated, i.e. $\rho_{X,Y} = 0$, the dissimilarity is $d(X,Y) = 1$. When $X$ and $Y$ are entirely negatively correlated, and $\rho_{X,Y} = -1$, we have $d(X,Y) = 2$. We apply MDS to  the sentiment-topic-space, reducing it to a two-dimensional representation using dissimiliarties $1-\rho_{X,Y}$. This embedding is depicted in Figure \ref{fig:MDS}. For improved visualisation, we represent each topic by a word cloud where the size of each word is related to the word density in the topic, $p(\text{word}|\text{topic})$.

Figure \ref{fig:MDS} shows clusters of sentiments. On the left is sadness, anger, negative, disgust, and fear, and on the right positive sentiments; trust, positive, and joy. This clustering also contains the sentiment surprise, which is not necessarily a positive emotion. Anticipation is located relatively far from all other sentiments. These clusters correspond to the highest levels of clustering indicated by the dendrogram in Figure \ref{fig:heatmap}. Topics are coloured to match that of the nearest sentiment in the two-dimensional embedding. We remark that the placement of topics near sentiments in Figure \ref{fig:MDS} does not imply that they necessarily pertain to that sentiment, merely, they correlate well with it.

\section{Limitations}

There are two main types of limitations that should be considered; those stemming from the data, and resulting from the methods used to analyse the data. 

First, we consider limitations in the data. Posts to \texttt{r/COVID19Positive} are subject to many biases. Notably, there is a bias in those who use Reddit.
In particular, while a global platform, its user base is largely North American \cite{gjurkovic2020pandora}, and so we should expect that these results are largely describing a North American experience of the virus.
It is also clear that, similar to other forms of social media such as Twitter \cite{mitchell2013geography}, there will also be significant demographic biases, on Reddit towards younger, more educated, politically-liberal, users \cite{barthel2016nearly}.
We also acknowledge there is a quite literal survivorship bias present. Clearly, people who have passed away due to COVID-19, and those in critical cases will be unable to post about their experiences. There is an avenue to explore these cases through the flair ``Tested Positive - Family'' that future research may consider.

In addition to biases, the quality of the content of these posts is not held to any standard, and posts are purely anecdotal. This is in contrast to traditional evidence-based medicine, which de-emphasises anecdotal reports \cite{charalambous2019social}. Moreover, there is no way to ensure authors posting under the flair ``Tested Positive - Me'' and ``Tested Positive'' did in fact test positive for COVID-19. There will also be inconsistencies to the day the authors define as Day 1. In this analysis we remark that this day is the first day of symptom onset, however, some authors may refer to Day 1 as the first day since testing positive. This would be especially true in the case of an asymptomatic person who tested positive. Together, these form considerable limitations towards the reliability of the narratives, and hence the resulting analysis. We, therefore, advise caution when interpreting this analysis, and that these limitations are kept in mind.
On the other hand, this remains an important corpus for study, partly due to its visibility -- that these narratives are publicly posted means they are potentially more influential in the broader media than truly personal diaries.

There are also limitations within the methods used. While significant effort was made during preprocessing to annotate texts by their reference date, it was not perfect. People often write with flashbacks to previous dates, producing incorrect annotations. Future research will attempt to capture this more effectively. People also tend to post a range of dates. It is conceivable that after topic modelling, we duplicate the results for each day in the range mentioned. However, as there is little data available, this would bias the results towards those posts that include ranges. Instead of incorporating the range, we take the midpoint of the range as the day given for the annotation.

Sentiment often changes with context. In one setting, a statement may be positive, and in others, the same statement may be negative. The sentiment analysis conducted in this paper was not context-specific. We provide sentiment analysis in a general context. As a result, medical terminology such as ``fever'', ``cough'', ``doctor'' contribute towards specific sentiments. In the NRC lexicon, ``doctor'', ``hospital'', and ``nurse'' possess the sentiment ``trust'' (Table \ref{tab:sentiment_words}). In the context of COVID-19 positive narratives, this may not be an indicator of trust, and instead, an indicator of a serious condition. Ideally, we would be relatively agnostic about our \textit{a priori} belief of the sentiment of words whose sentiment may be subject to change in different contexts. Future work will look at combined sentiment-topic modelling approaches to attempt to incorporate context more effectively.

As time passes and more people document their experiences, this dataset will grow, and so too does the potential to extract more detailed topics, and in general conduct a more sophisticated analysis. We will continue to monitor \texttt{r/COVIDpositive} over time to analyse longer-term trends in the way patients describe their experience with the virus.

\section{Conclusion}

We analysed personal narratives of being COVID-19 positive through a collective timeline using topic modelling and sentiment analysis. Topic modelling revealed clusters of related symptoms in topics. The densities of these topics are presented across the collective timeline. Topics regarding early-onset symptoms, containing terms such as ``fever'', ``cough'', ``throat'', and ``nose'' are most prominent towards the beginning of the timeline. Topics containing terms such as ``breathing'', ``chest'', ``lungs'', indicating more serious conditions, appear more frequently in later stages of the timeline. Sentiment analysis revealed a high level of negativity and fear in these narratives. The relationships between topics and sentiments are explored through their Pearson correlations. Hierarchical clustering and MDS bring forth structure in these relationships. As a result, we observe further groupings of early symptoms than found by topic modelling alone, as well as groupings of sentiments with these topics.

The approaches employed in this analysis have demonstrated the ability to capture data that can be linked to rapidly changing events, such as the COVID-19 pandemic. Social media sources, such as Reddit, provide a source of data not traditionally recorded for medical analysis, that can be used to detect the sentiment of a population, and also identify the changing influence of symptoms over time. Sentiment may be important in highlighting the need for other supports, such as mental health support, stemming from the pandemic.

This paper identifies areas of intended future research. Looking beyond the flairs indicating the author of the post has tested positive, to those that include family members testing positive, and questions people pose to those COVID-19 positive will provide a deeper understanding of how COVID-19 has an indirect effect on people through their families, friends, and concerns. The development of more sophisticated, context specific sentiment analysis will better capture how we feel towards COVID-19 and how it affects our mental health. Improvements to the data preprocessing such as annotation of dates will improve the signal-to-noise ratio, and yield a more accurate and reliable analysis. Similarly, the collection of future narratives yet to be told will improve the sensitivity of the analysis, and allow further techniques to be used. Furthermore, the assimilation of data relating to this pandemic with future epidemics and/or pandemics will permit similar analyses comparing social media responses to be conducted.

\bibliography{bibliography}

\begin{thebibliography}{27}
\providecommand{\natexlab}[1]{#1}
\providecommand{\url}[1]{\texttt{#1}}
\expandafter\ifx\csname urlstyle\endcsname\relax
  \providecommand{\doi}[1]{doi: #1}\else
  \providecommand{\doi}{doi: \begingroup \urlstyle{rm}\Url}\fi

\bibitem[Ordun et~al.(2020)Ordun, Purushotham, and Raff]{ordun2020exploratory}
Catherine Ordun, Sanjay Purushotham, and Edward Raff.
\newblock Exploratory analysis of {COVID}-19 tweets using topic modeling,
  {UMAP, and DiGraphs}.
\newblock \emph{arXiv preprint arXiv:2005.03082}, 2020.

\bibitem[Sha et~al.(2020)Sha, Hasan, Mohler, and Brantingham]{sha2020dynamic}
Hao Sha, Mohammad~Al Hasan, George Mohler, and P~Jeffrey Brantingham.
\newblock Dynamic topic modeling of the {COVID-19} {T}witter narrative among
  {US} governors and cabinet executives.
\newblock \emph{arXiv preprint arXiv:2004.11692}, 2020.

\bibitem[Chen et~al.(2020{\natexlab{a}})Chen, Lyu, Yang, Wang, and
  Luo]{chen2020eyes}
Long Chen, Hanjia Lyu, Tongyu Yang, Yu~Wang, and Jiebo Luo.
\newblock In the eyes of the beholder: Sentiment and topic analyses on social
  media use of neutral and controversial terms for {COVID-19}.
\newblock \emph{arXiv preprint arXiv:2004.10225}, 2020{\natexlab{a}}.

\bibitem[Li et~al.(2020)Li, Li, Li, Alvarez-Napagao, and Garcia]{li2020we}
Irene Li, Yixin Li, Tianxiao Li, Sergio Alvarez-Napagao, and Dario Garcia.
\newblock What are we depressed about when we talk about {COVID19}: Mental
  health analysis on tweets using natural language processing.
\newblock \emph{arXiv preprint arXiv:2004.10899}, 2020.

\bibitem[Wicke and Bolognesi(2020)]{wicke2020framing}
Philipp Wicke and Marianna~M Bolognesi.
\newblock Framing {COVID-19}: How we conceptualize and discuss the pandemic on
  {T}witter.
\newblock \emph{arXiv preprint arXiv:2004.06986}, 2020.

\bibitem[Reagan et~al.(2016)Reagan, Mitchell, Kiley, Danforth, and
  Dodds]{reagan2016emotional}
Andrew~J Reagan, Lewis Mitchell, Dilan Kiley, Christopher~M Danforth, and
  Peter~Sheridan Dodds.
\newblock The emotional arcs of stories are dominated by six basic shapes.
\newblock \emph{EPJ Data Science}, 5\penalty0 (1):\penalty0 31, 2016.

\bibitem[Alajajian et~al.(2017)Alajajian, Williams, Reagan, Alajajian, Frank,
  Mitchell, Lahne, Danforth, and Dodds]{alajajian2017lexicocalorimeter}
Sharon~E Alajajian, Jake~Ryland Williams, Andrew~J Reagan, Stephen~C Alajajian,
  Morgan~R Frank, Lewis Mitchell, Jacob Lahne, Christopher~M Danforth, and
  Peter~Sheridan Dodds.
\newblock The {L}exicocalorimeter: {G}auging public health through caloric
  input and output on social media.
\newblock \emph{PloS one}, 12\penalty0 (2), 2017.

\bibitem[Reece et~al.(2017)Reece, Reagan, Lix, Dodds, Danforth, and
  Langer]{reece2017forecasting}
Andrew~G Reece, Andrew~J Reagan, Katharina~LM Lix, Peter~Sheridan Dodds,
  Christopher~M Danforth, and Ellen~J Langer.
\newblock Forecasting the onset and course of mental illness with {T}witter
  data.
\newblock \emph{Scientific reports}, 7\penalty0 (1):\penalty0 1--11, 2017.

\bibitem[Tiggemann et~al.(2018)Tiggemann, Churches, Mitchell, and
  Brown]{tiggemann2018tweeting}
Marika Tiggemann, Owen Churches, Lewis Mitchell, and Zoe Brown.
\newblock Tweeting weight loss: {A} comparison of \#thinspiration and
  \#fitspiration communities on twitter.
\newblock \emph{Body Image}, 25:\penalty0 133--138, 2018.

\bibitem[Hawn(2009)]{hawn2009take}
Carleen Hawn.
\newblock Take two aspirin and tweet me in the morning: how {T}witter,
  {F}acebook, and other social media are reshaping health care.
\newblock \emph{Health affairs}, 28\penalty0 (2):\penalty0 361--368, 2009.

\bibitem[O'Connor(2017)]{o2017using}
Siobhan O'Connor.
\newblock Using social media to engage nurses in health policy development.
\newblock \emph{Journal of nursing management}, 25\penalty0 (8):\penalty0
  632--639, 2017.

\bibitem[Charalambous(2019)]{charalambous2019social}
Andreas Charalambous.
\newblock Social media and health policy.
\newblock \emph{Asia-Pacific journal of oncology nursing}, 6\penalty0
  (1):\penalty0 24, 2019.

\bibitem[Roland(2018)]{roland2018social}
Damian Roland.
\newblock Social media, health policy, and knowledge translation.
\newblock \emph{Journal of the American College of Radiology}, 15\penalty0
  (1):\penalty0 149--152, 2018.

\bibitem[Bou-Karroum et~al.(2017)Bou-Karroum, El-Jardali, Hemadi, Faraj, Ojha,
  Shahrour, Darzi, Ali, Doumit, Langlois, et~al.]{bou2017using}
Lama Bou-Karroum, Fadi El-Jardali, Nour Hemadi, Yasmine Faraj, Utkarsh Ojha,
  Maher Shahrour, Andrea Darzi, Maha Ali, Carine Doumit, Etienne~V Langlois,
  et~al.
\newblock Using media to impact health policy-making: an integrative systematic
  review.
\newblock \emph{Implementation Science}, 12\penalty0 (1):\penalty0 52, 2017.

\bibitem[Shaw et~al.(2020)Shaw, Kim, and Hua]{shaw2020governance}
Rajib Shaw, Yong-kyun Kim, and Jinling Hua.
\newblock Governance, technology and citizen behavior in pandemic: Lessons from
  {COVID-19} in {East Asia}.
\newblock \emph{Progress in Disaster Science}, page 100090, 2020.

\bibitem[Gerlach et~al.(2018)Gerlach, Peixoto, and Altmann]{gerlach2018network}
Martin Gerlach, Tiago~P Peixoto, and Eduardo~G Altmann.
\newblock A network approach to topic models.
\newblock \emph{Science advances}, 4\penalty0 (7):\penalty0 eaaq1360, 2018.

\bibitem[Blei et~al.(2003)Blei, Ng, and Jordan]{blei2003latent}
David~M Blei, Andrew~Y Ng, and Michael~I Jordan.
\newblock Latent {D}irichlet allocation.
\newblock \emph{Journal of machine Learning research}, 3\penalty0
  (Jan):\penalty0 993--1022, 2003.

\bibitem[Peixoto(2014)]{peixoto2014hierarchical}
Tiago~P Peixoto.
\newblock Hierarchical block structures and high-resolution model selection in
  large networks.
\newblock \emph{Physical Review X}, 4\penalty0 (1):\penalty0 011047, 2014.

\bibitem[Peixoto(2015)]{peixoto2015model}
Tiago~P Peixoto.
\newblock Model selection and hypothesis testing for large-scale network models
  with overlapping groups.
\newblock \emph{Physical Review X}, 5\penalty0 (1):\penalty0 011033, 2015.

\bibitem[Peixoto(2017)]{peixoto2017nonparametric}
Tiago~P Peixoto.
\newblock Nonparametric {B}ayesian inference of the microcanonical stochastic
  block model.
\newblock \emph{Physical Review E}, 95\penalty0 (1):\penalty0 012317, 2017.

\bibitem[Chen et~al.(2020{\natexlab{b}})Chen, Qi, Liu, Ling, Qian, Li, Li, Xu,
  Zhang, Xu, et~al.]{chen2020clinical}
Jun Chen, Tangkai Qi, Li~Liu, Yun Ling, Zhiping Qian, Tao Li, Feng Li, Qingnian
  Xu, Yuyi Zhang, Shuibao Xu, et~al.
\newblock Clinical progression of patients with {COVID-19} in {S}hanghai,
  {C}hina.
\newblock \emph{Journal of Infection}, 2020{\natexlab{b}}.

\bibitem[Mao et~al.(2020)Mao, Wang, Chen, He, Chang, Hong, Zhou, Wang, Miao,
  Hu, et~al.]{mao2020neurological}
Ling Mao, Mengdie Wang, Shengcai Chen, Quanwei He, Jiang Chang, Candong Hong,
  Yifan Zhou, David Wang, Xiaoping Miao, Yu~Hu, et~al.
\newblock Neurological manifestations of hospitalized patients with {COVID-19
  in Wuhan, C}hina: a retrospective case series study.
\newblock 2020.

\bibitem[Mohammad and Turney(2013)]{mohammad13}
Saif~M. Mohammad and Peter~D. Turney.
\newblock Crowdsourcing a word-emotion association lexicon.
\newblock \emph{Computational Intelligence}, 29\penalty0 (3):\penalty0
  436--465, 2013.
\newblock \doi{10.1111/j.1467-8640.2012.00460.x}.
\newblock URL
  \url{https://onlinelibrary.wiley.com/doi/abs/10.1111/j.1467-8640.2012.00460.x}.

\bibitem[Plutchik and Kellerman(1980)]{plutchik1980emotion}
Robert Plutchik and Henry Kellerman.
\newblock \emph{Emotion, theory, research, and experience}.
\newblock Academic press, 1980.

\bibitem[Gjurkovi{\'c} et~al.(2020)Gjurkovi{\'c}, Karan, Vukojevi{\'c},
  Bo{\v{s}}njak, and {\v{S}}najder]{gjurkovic2020pandora}
Matej Gjurkovi{\'c}, Mladen Karan, Iva Vukojevi{\'c}, Mihaela Bo{\v{s}}njak,
  and Jan {\v{S}}najder.
\newblock {PANDORA} talks: Personality and demographics on {R}eddit.
\newblock \emph{arXiv preprint arXiv:2004.04460}, 2020.

\bibitem[Mitchell et~al.(2013)Mitchell, Frank, Harris, Dodds, and
  Danforth]{mitchell2013geography}
Lewis Mitchell, Morgan~R Frank, Kameron~Decker Harris, Peter~Sheridan Dodds,
  and Christopher~M Danforth.
\newblock The geography of happiness: Connecting twitter sentiment and
  expression, demographics, and objective characteristics of place.
\newblock \emph{PloS one}, 8\penalty0 (5), 2013.

\bibitem[Barthel et~al.(2016)Barthel, Stocking, Holcomb, and
  Mitchell]{barthel2016nearly}
Michael Barthel, Galen Stocking, Jesse Holcomb, and Amy Mitchell.
\newblock Nearly eight-in-ten reddit users get news on the site.
\newblock \emph{Pew Research Center}, 25, 2016.

\end{thebibliography}
\end{document}